# TOPOLOGY DESIGN AND ANALYSIS OF A NOVEL 3-TRANSLATIONAL PARALLEL MECHANISM WITH ANALYTICAL DIRECT POSITION SOLUTIONS AND PARTIAL MOTION DECOUPLING

**Boxiong Zeng**

School of Mechanical Engineering, Changzhou University, Changzhou 213016, China
710022180@qq.com

**Ting-li Yang**

School of Mechanical Engineering, Changzhou University, Changzhou 213016, China
yangtl@126.com

**Huiping Shen**

School of Mechanical Engineering, Changzhou University, Changzhou 213016, China
shp65@126.com

**Damien Chablat**

CNRS, Laboratoire des Sciences du Numérique de Nantes, UMR 6004 Nantes, France

Damien.Chablat@cnrs.fr

**ABSTRACT**

*According to the topological design theory and method of parallel mechanism (PM) based on position and orientation characteristic (POC) equations, this paper design a novel 3-translation (3T) PM that has three advantages, i.e., ① it consists on three actuated prismatic joints, ② the PM has analytical direct position solutions, and ③ the PM is of partial motion decoupling property. Firstly, the main topological characteristics such as the POC, degree of freedom and coupling degree are calculated for kinematics modelling. Due to the special constraint feature of the 3-translation, the analytical direct position solutions of the PM can be directly obtained without needing to use one-dimensional search method. Further, the conditions of the singular configuration of the PM, as well as the singularity location inside the workspace are analyzed according to the inverse kinematics.*

**INTRODUCTION**

In many industrial production lines, process operations require pure translation movements only. Therefore, the 3-DOF translational parallel mechanism (TPM) has a significant potential value due to its small number of actuated components, relatively simple structure and easily to be controlled.

Many scholars have being studied the TPM. For example, original design of 3-DOF TPM is the Delta Robot which was presented by Clavel [1]. The Delta-based structure manipulators of TPM have been developed [2-4]. Tsai et al [5, 6] presented the 3-DOF TPM, the moving actuators of which are prismatic joints and the sub-chain is 4R parallelogram mechanism (P is prismatic joint and R is revolute joint). In [7, 8], the authors suggested a 3-RRC TPM and developed the kinematics and workspace analysis (C is cylindrical joint). Kong et al [9] proposed a 3-CRR mechanism with good motion performance and no obvious singular position. Li et al [10, 11] developed a 3-UPU PM (U is universal joint) and analyzed the instantaneous kinematics performance of the TPM. Yu et al [12] carried out a comprehensive analysis of the three-dimensional TPM configuration based on the screw theory. Lu et al [13] proposed a 3-RRRP (4R) three-translation PM and analyzed the kinematics and workspace. Yang et al [14-17] based on the single opened chains (SOC) units theory to synthesize the 3T0R PM, a variety of new TPMs were synthesized and then classified, Considering the anisotropy of kinematics, Zhao et al [18] analyzed the dimension synthesis and kinematics of the 3-DOF TPM based on the Delta PM. Zeng et al [19-21] introduced a 3-DOF TPM called as Tri-pyramid robot and presented a more detailed analytical approach for the Jacobi matrix. Prause et al [22] compared the characteristics of dimensional synthesis, boundary conditions and workspace for various 3-DOF TPM for the better performance. Mahmood et al [23] proposed a 3-DOF 3-[P2(US) mechanism and analyzed its kinematics and dexterity.

However, the most previous TPMs generally suffer from two major problems: i) the degree of coupling κ of these PMs is greater than zero, which means its direct position solution is generally not analytical, and ii) these PMs do not have input-output decoupling characteristics [24], which leads to the complexity of motion control and path planning.

According to topology design theory of PM based on position and orientation characteristics (POC) equations [15, 18], a new TPM is proposed in this paper. The TPM is designed with simple structure and features one coupling degree, it consists of prismatic and revolute joints, and has analytical direct position solutions and partial motion decoupling property. The position solutions, singularity, workspace and its internal singularity of the PM are analyzed.

## DESIGN AND TOPOLOGY ANALYSIS

### Topological design

The 3T parallel manipulator proposed in this paper is illustrated in Fig. 1. The base platform 0 is connected to the moving platform 1 by two hybrid chains that contain looped Simple-Opened-Chains. Such hybrid chains are called HSOCs, their structural and geometric constraints are as follows:

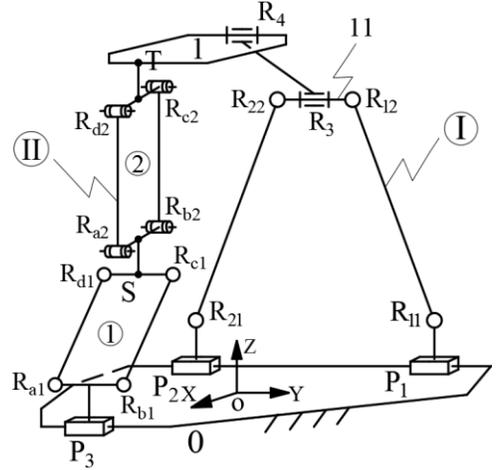

**FIGURE 1 KINEMATIC STRUCTURE OF THE 3T PM**

(1) For the 6-bar planar mechanism loop (abbreviation: 2P4R planar mechanism) in right side of Fig. 1, two revolute joints $R_3$ and $R_4$ whose axes are parallel to each other are connected in series, where $R_3$ is connected to link 11 and $R_4$ is connected to the moving platform 1 to obtain the first HSOC branch (denoted as: hybrid chain I). Two prismatic joints $P_1$ and $P_2$ will be used to be actuated.

(2) The left side branch is made up of a prismatic joint $P_3$ and two 4R parallelogram mechanisms connected in series, and the parallelograms connected from $P_3$ to the moving platform 1 are respectively recorded as ①, ②, after $P_3$ and the parallelogram ① are rigidly connected in the same plane, they are connected to the parallelogram ② in their orthogonal plane to obtain the second HSOC branch (denoted as: hybrid chain II).

(3) The prismatic joints $P_1$, $P_2$ and $P_3$ are connected to the base platform 0; $P_1$ and $P_2$ are arranged coaxially, and prismatic $P_1$ is parallel to $P_2$. When the PM is moving, the 2P4R planar mechanism is always parallel to the plane of the parallelogram ①.

### Analysis of topology characteristics

*Analysis of the POC set*

The POC set equations for serial and parallel mechanisms are expressed respectively as follows:

$$M_{bi} = \bigcup_{i=1}^{m} M_{Ji} \qquad (1)$$

$$M_{Pa} = \bigcap_{i=1}^{n} M_{bi} \qquad (2)$$

where

$M_{Ji}$ - POC set generated by the $i$-th joint.

$M_{bi}$ - POC set generated by the end link of $i$-th branched chain.

$M_{Pa}$ - POC set generated by the moving platform of PM.

Obviously, the output motion of the intermediate link 11 of the 2P4R planar mechanism on the hybrid chain I is two translations and one rotation (2T1R). The output motion of the link S of the parallelogram ① on the hybrid chain II is two translations (2T), and the output motion of the link T of the parallelogram ② is three translations (3T). Therefore, the topological architecture of the hybrid chain I and II of the PM can be equivalently denoted as, respectively:

$$HSOC_1\{-\Diamond(P_1^{(2P4R)}, P_2^{(2P4R)}) \perp R^{(2P4R)} \perp R_3 \parallel R_4 -\}$$

$$HSOC_2\{-P_3 - P^{(4R)} - P^{(4R)} -\}$$

The POC sets of the end of the two HSOCs are determined according to Eqs. (1) and (2) as follows:

$$M_{HSOC_1} = \begin{bmatrix} t^2(\perp R_{12}) \\ r^1(\parallel R_{12}) \end{bmatrix} Y \begin{bmatrix} t^2(\perp R_3) \\ r^1(\parallel R_3) \end{bmatrix} = \begin{bmatrix} t^3 \\ r^2(\parallel \Diamond(R_{12}, R_3)) \end{bmatrix}$$

$$M_{HSOC_2} = \begin{bmatrix} t^1(\parallel P_3) \\ r^0 \end{bmatrix} \underset{i=1}{\overset{2}{Y}} \begin{bmatrix} t^1(\parallel \Diamond(R_{ai}R_{bi}R_{ci}R_{di})) \\ r^0 \end{bmatrix} = \begin{bmatrix} t^3 \\ r^0 \end{bmatrix}$$

The POC set of the moving platform of this PM is determined from Eq. (2) by

$$M_{Pa} = M_{HSOC_1} \cap M_{HSOC_2} = \begin{bmatrix} t^3 \\ r^0 \end{bmatrix}$$

This formula indicates that the moving platform 1 of the PM produces three translations motion. It is further known that the hybrid chain II in the mechanism itself can realize the design requirement of three translations, which simultaneously constrains the two rotational outputs of the hybrid chain I.

*Determining the DOF*

The general and full-cycle DOF formula for PMs proposed in author's work [16] is given below:

$$F = \sum_{i=1}^{m} f_i - \sum_{j=1}^{v} \xi_{Lj} \qquad (3)$$

$$\sum_{j=1}^{v} \xi_{Lj} = \dim \left\{ \left( \bigcap_{i=1}^{j} M_{b_i} \right) Y M_{b_{(j+1)}} \right\} \qquad (4)$$

where

$F$ - DOF of PM.

$f_i$ - DOF of the $i^{th}$ joint.

$m$ - number of all joints of the PM.

$v$ - number of independent loops, and $v=m-n+1$.

$n$ - number of links.

$\xi_{L_j}$ - number of independent equations of the $j^{th}$ loop.

$\bigcap_{i=1}^{j} M_{b_i}$ - POC set generated by the sub-PM formed by the former $j$ branches.

$M_{b(j+1)}$ - POC set generated by the end link of $j+1$ sub-chains.

The PM can be decomposed into two independent loops, and their constraint equations are calculated as follows:

①The first independent loop is consisted of the 2P4R planar mechanism in the hybrid chain I, the $LOOP_1$ is deduced as:

$$LOOP_1\{-\Diamond(P_1^{(2P4R)}, P_2^{(2P4R)}) \perp R^{(2P4R)} -\}$$

Obviously, the independent displacement equation number $\xi_{L_1} = 3$.

②The above 2P4R planar mechanism and the following sub-string $R_3 \parallel R_4$ plus $HSOC_2$ will form the second independent loop, that is

$$LOOP_2\{-R_3 \parallel R_4 - P^{(4R)} - P^{(4R)} - P_3 -\}$$

In accordance with Eq.(4), the independent displacement equation number $\xi_{L_2}$ of the second loop can be obtained as follows:

$$\xi_{L_2} = \dim \left\{ \begin{bmatrix} t^2(\perp R_{12}) \\ r^1(\parallel R_{12}) \end{bmatrix} Y \begin{bmatrix} t^3 \\ r^1(\parallel R_3) \end{bmatrix} \right\} = \dim \left\{ \begin{bmatrix} t^3 \\ r^2(\parallel \Diamond(R_{12}, R_3)) \end{bmatrix} \right\} = 5$$

Thus, the DOF of the PM is calculated from Eq. (3) as

$$F = \sum_{i=1}^{m} f_i - \sum_{j=1}^{2} \xi_{L_j} = (6+5) - (3+5) = 3$$

Therefore, the DOF of the PM is 3, and when the prismatic joints $P_1$, $P_2$ and $P_3$ on the base platform 0 are the actuated joints, the moving platform 1 can realize 3-translational motion outputs.

*Determining the coupling degree*

According to the composition principle of mechanism based on single-opened-chains (SOC) units, any PM can be decomposed into a series of Assur kinematics chains (AKC), and an AKC with $v$ independent loops can be decomposed into $v$ SOC. The constraint of the $j^{th}$ SOC is defined [16,17] by

$$\Delta_j = \sum_{i=1}^{m_j} f_i - I_j - \xi_{L_j} = \begin{cases} \Delta_j^- = -5,-4,-2,-1 \\ \Delta_j^0 = 0 \\ \Delta_j^+ = +1,+2,+3,\cdots \end{cases} \quad (5)$$

where

$m_j$ - number of joints contained in the $j^{th}$ SOC$_j$.

$f_i$ - DOF of the $i^{th}$ joints.

$I_j$ - number of actuated joints in the $j^{th}$ SOC$_j$.

$\xi_{L_j}$ - number of independent equations of the $j^{th}$ loop.

For an AKC, it must be satisfied

$$\sum_{j=1}^{v} \Delta_j = 0$$

then, the coupling degree of AKC [16,17] is

$$\kappa = \frac{1}{2} min \left\{ \sum_{j=1}^{v} |\Delta_j| \right\} \quad (6)$$

The physical meaning of the coupling degree $\kappa$ can be explained in this way. The coupling degree $\kappa$ reflects the correlation and dependence between kinematic variables of each independent loop of the mechanism. It has been proved that the higher $\kappa$, the greater the complexity of the kinematic and dynamic solutions of the mechanism will be.

The independent displacement equations of $LOOP_1$ and $LOOP_2$ have been calculated in the previous section

*Determining the DOF*, i.e., $\xi_{L_1} = 3$, $\xi_{L_2} = 5$, thus, the degree of constraint of the two independent loops are calculated by Eq. (5), respectively, can be obtained as follows:

$$\Delta_1 = \sum_{i=1}^{m_1} f_i - I_1 - \xi_{L_1} = 6 - 2 - 3 = 1$$

$$\Delta_2 = \sum_{i=1}^{m_2} f_i - I_2 - \xi_{L_2} = 5 - 1 - 5 = -1$$

The coupling degrees of the AKC is calculated by Eq. (6) as

$$k = \frac{1}{2} \sum_{j=1}^{v} |\Delta_j| = \frac{1}{2}(|+1| + |-1|) = 1$$

Thus, the PM contains only one AKC, and its coupling degrees equals to 1. Therefore, when solving the direct position solutions of the PM, it is necessary to set only one virtual variable in the loop whose degree of constraint is positive $(\Delta_j > 0)$. Then, a position constraint equation with this virtual variable is established in the loop with the negative constraint $(\Delta_j < 0)$, and the real value of the virtual variable can be obtained by the one-dimensional search method, accordingly to obtain the direct position solutions of the PM.

However, owing to the special 3-translational feature constraint of the PM, the loop with the negative degree of constraint $(\Delta_j < 0)$ can be directly applied to the geometric constraint of the loop with the positive degree of constraint $(\Delta_j > 0)$, i.e., The motion of the link 11 is always parallel to the base platform 0, from which the virtual variable is easily obtained, and there is no need to solve the virtual variable by one-dimensional search method. Therefore, the analytical direct position solutions of the PM can be directly obtained in the following section, which greatly simplifies the process of the direct solutions. This method of solving the direct solutions of the virtual variable directly by special geometric constraints is of general.

## POSITION ANALYSIS

### The coordinate system and parameterization

The kinematics modeling of the PM is shown in Fig. 2. The base platform 0 is a rectangle having a length and a width of $2a$ and $2b$ respectively. The frame coordinate system O-XYZ is established on the geometric center of the base platform 0, the X and Y axes of which are perpendicular and parallel to the line $A_1A_2$, and the Z axis is determined by the right hand Cartesian coordinate rule. The moving coordinate system O′-X′Y′Z′ is established at the center of the moving platform 1, the X′ and Y′ axes of which are coincide and perpendicular to the line $D_2F_3$, and the Z′ axis is determined the right hand Cartesian coordinate rule.

The length of the three driving links 2 is $l_1$, the length of the connecting links 9 and 10 on the hybrid chain I is $l_2$, and the lengths of the intermediate links 11 and 12 are $l_3$, $l_4$ respectively.

The length of the parallelogram short links 3, 6 on the hybrid chain II is $l_5$. the point $B_3$, $C_3$, $D_3$ and $E_3$ are the midpoint of the short edge, the length of the long links 4, 7 is $l_6$, The length of the connecting link 5 between the parallelograms is $l_7$. The length of the connecting link 8 is $l_8$, and the length of the line $D_2F_3$ on the moving platform 1 is $2d$.

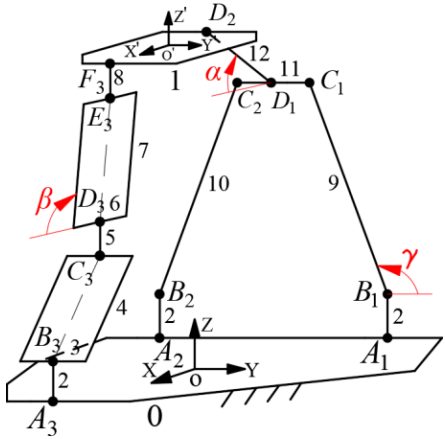

**(A) KINEMATIC MODELING**

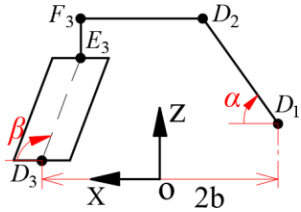

**(B) GEOMETRIC RELATIONSHIP OF THE SECOND LOOP (PARTIAL) IN THE XOZ DIRECTION**

**FIGURE 2 KINEMATIC MODELING OF THE 3T PM**

The angle between the vectors $B_1C_1$ and the Y axis is $\gamma$, and the $\gamma$ is assigned as virtual variable. The angles between the vectors $D_1D_2$, $D_3E_3$ and the X axis are $\alpha$ and $\beta$ respectively.

**Direct kinematics**

To solve the direct kinematics, i.e., to compute the position $O'(x,y,z)$ of the moving platform when setting the position values of the prismatic joints $P_1$, $P_2$ and $P_3$ (with the coordinates $y_{A_1}$, $y_{A_2}$ and $y_{A_3}$).

**1）Solving the first loop with positive degree of constraint**

$$LOOP_1: A_1 - B_1 - C_1 - C_2 - B_2 - A_2$$

The coordinates of points $A_1$, $A_2$ and $A_3$ on the base platform 0 are respectively

$$A_1 = (-b, y_{A_1}, 0)^T, A_2 = (-b, y_{A_2}, 0)^T, A_3 = (b, y_{A_3}, 0)^T.$$

The coordinates of each end-point of the three same actuated links 2, i.e., $B_1$, $B_2$ and $B_3$ are easily calculated as

$$B_1 = (-b, y_{A_1}, l_1)^T, B_2 = (-b, y_{A_2}, l_1)^T, B_3 = (b, y_{A_3}, l_1)^T.$$

Due to the special constraint of the three translations of the moving platform 1, during the movement of the PM, the intermediate link 11 of the 2P4R planar mechanism is always parallel to the base platform 0, that is, $C_1C_2 \| A_1A_2$, then we have

$$z_{C_1} = z_{C_2} \qquad (7)$$

Therefore, the coordinates of points $C_1$ and $C_2$ are calculated as

$$C_1 = (-b, y_{A_1} + l_2\cos\gamma, l_1 + l_2\sin\gamma)^T$$

$$C_2 = (-b, y_{A_1} + l_2\cos\gamma - l_3, l_1 + l_2\sin\gamma)^T$$

Due to the link length constraints defined by $B_2C_2 = l_2$, the constraint equation can be deduced as below.

$$(x_{C_2} - x_{B_2})^2 + (y_{C_2} - y_{B_2})^2 + (z_{C_2} - z_{B_2})^2 = l_2^2 \qquad (8)$$

Eq. (8) leads to

$$AB\cos\gamma + B^2 = 0$$

① when $B=0$, the value of $\gamma$ cannot be determined at this time, and the PM has parallel singularity, this situation of which should be avoided.

② when $B \neq 0$, the value of $\gamma$ can be determined at this time, there is

$$\gamma = \arccos \frac{\pm B}{A} \quad (9)$$

where

$$A = 2l_2, B = y_{A_1} - l_3 - y_{A_2}.$$

Thus, the second loop with negative degree of constraint acts on the special geometric constraint Eq. (7) on the first loop with positive degree of constraint, which is the key to directly finding the analytical solutions of the virtual variable $\gamma$. This is an advantage for easy obtaining the analytical direct solutions from the topological constraint analysis of the PM.

**2) Solving the second loop with negative degree of constraint**

$$LOOP_2 : D_1 - D_2 - F_3 - E_3 - D_3 - C_3 - B_3 - A_3$$

The coordinates of points $D_1$ and $D_2$ obtained from points $C_1$ and $C_2$ are calculated as

$$D_1 = (-b, y_{A_1} + l_2 \cos\gamma - l_3/2, l_1 + l_2 \sin\gamma)^T$$

$$D_2 = \begin{bmatrix} -b + l_4 \cos\alpha \\ y_{A_1} + l_2 \cos\gamma - l_3/2 \\ l_1 + l_2 \sin\gamma + l_4 \sin\alpha \end{bmatrix}$$

Simultaneously, the coordinates of point $O'$ can be calculated as:

$$O' = \begin{bmatrix} x \\ y \\ z \end{bmatrix} = \begin{bmatrix} -b + l_4 \cos\alpha + d \\ y_{A_1} + l_2 \cos\gamma - l_3/2 \\ l_1 + l_2 \sin\gamma + l_4 \sin\alpha \end{bmatrix} \quad (10)$$

Further, the coordinates of points $F_3$, $E_3$, $D_3$ and $C_3$ are represented by the coordinates of point $O'$ defined as:

$$F_3 = (x+d, y, z)^T$$

$$E_3 = (x+d, y, z-l_8)^T$$

$$D_3 = (b, y, z-l_8 - l_6 \sin\beta)^T$$

$$C_3 = (b, y, z-l_8 - l_6 \sin\beta - l_7)^T \quad (11)$$

Due to the link length constraints defined by $B_3C_3 = l_6$, the constraint equation can be deduced as below.

$$(x_{C_3} - x_{B_3})^2 + (y_{C_3} - y_{B_3})^2 + (z_{C_3} - z_{B_3})^2 = l_6^2 \quad (12)$$

and let

$$l_4 \sin\alpha - l_6 \sin\beta = t \quad (13)$$

then, there is

$$(H_1 + t)^2 - H_2 = 0$$

$$t = -H_1 \pm \sqrt{H_2} \quad (14)$$

where

$$H_1 = l_2 \sin\gamma - l_8 - l_7, \ H_2 = l_6^2 - (y_{C_3} - y_{B_3})^2.$$

When the PM is moving, the 2P4R planar mechanism is always parallel with the plane of the parallelogram ①, therefore, there is always relation as

$$y_{D_1} = y_{D_3} \quad (15)$$

$$l_4 \cos\alpha + 2d - l_6 \cos\beta = 2b \quad (16)$$

After eliminate $\beta$ from Eqs. (13) and (16), there is

$$J_1 \sin\alpha + J_2 \cos\alpha + J_3 = 0$$

let $k_1 = \tan\frac{\alpha}{2}$

$$\alpha = 2\arctan\frac{J_1 \pm \sqrt{J_1^2 + J_2^2 - J_3^2}}{J_2 - J_3} \quad (17)$$

where

$$J_1 = 2l_4 t, J_2 = 4l_4(b-d),$$
$$J_3 = l_6^2 - l_4^2 - t^2 - 4(b-d)^2.$$

Finally, putting the values of $\gamma$ and $\alpha$ obtained by Eqs. (9) and (17) respectively, into Eq. (10), the coordinates of point

$O'$ on the moving platform 1 can be obtained.

From Eq. (9), $\gamma = f_1(y_{A_1}, y_{A_2})$.

From Eq. (17), $\alpha = f_2(y_{A_1}, y_{A_2}, y_{A_3})$.

Therefore, it is known from Eq. (10) that

$$\begin{cases} x = f_1'(y_{A_1}, y_{A_2}, y_{A_3}) \\ y = f_2'(y_{A_1}, y_{A_2}) \\ z = f_3'(y_{A_1}, y_{A_2}, y_{A_3}) \end{cases}$$

I.e., the PM has partial input-output motion decoupling, which is advantageous for trajectory planning and motion control of the moving platform.

For the convenience of understanding, the above calculation process can be described by Fig. 3.

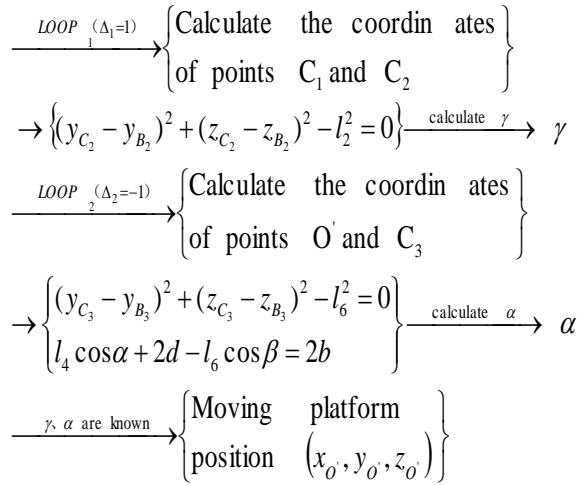

**FIGURE 3 FLOW CHART OF DIRECT POSITION SOLUTIONS**

It can be seen that the geometric constraint Eqs. (7), (15) and (16) are the key to find the analytical formula of the first and second loop position equations of the PM.

**Inverse kinematics**

To solve the inverse kinematics, we compute the values of $y_{A_1}$, $y_{A_2}$ and $y_{A_3}$ as a function of the coordinate $O'(x,y,z)$ of the moving platform.

For a given position of the moving platform, from Eqs. (10) and (16), the angles $\alpha$ and $\beta$ are calculated as

$$\alpha = \pm \arccos \frac{x+b-d}{l_4} \quad (18)$$

$$\beta = \pm \arccos \frac{x+d-b}{l_6} \quad (19)$$

Further, the coordinates of points $C_1$ and $C_2$ are defined as:

$$C_1 = (-b, y+l_3/2, z-l_4 \sin \alpha)^T$$

$$C_2 = (-b, y-l_3/2, z-l_4 \sin \alpha)^T$$

In addition, the coordinates of point $C_3$ have been given by Eq. (11). Therefore, due to the link length constraints defined by $B_1C_1 = B_2C_2 = l_2$ and $B_3C_3 = l_6$, there are three constraint equations as below.

$$\begin{cases} (x_{C_1} - x_{B_1})^2 + (y_{C_1} - y_{B_1})^2 + (z_{C_1} - z_{B_1})^2 = l_2^2 \\ (x_{C_2} - x_{B_2})^2 + (y_{C_2} - y_{B_2})^2 + (z_{C_2} - z_{B_2})^2 = l_2^2 \\ (x_{C_3} - x_{B_3})^2 + (y_{C_3} - y_{B_3})^2 + (z_{C_3} - z_{B_3})^2 = l_6^2 \end{cases} \quad (20)$$

From Eqs. (20), we can evaluate $y_{A_i}$ ($i=1,2,3$) as following

$$y_{A_i} = y_{C_i} \pm \sqrt{M_i} \quad (i=1,2,3) \quad (21)$$

where

$$M_1 = l_2^2 - (z_{C_1} - l_1)^2, M_2 = l_2^2 - (z_{C_2} - l_1)^2,$$

$$M_3 = l_6^2 - (z_{C_3} - l_1)^2.$$

In summary, when the coordinates of point $O'$ on the moving platform 1 are known, each input values $y_{A_1}$, $y_{A_2}$ and $y_{A_3}$ has two sets of solutions. Therefore, the inverse solution number is $2 \times 2 \times 8 = 32$.

**Numerical valuation for direct and inverse solutions**

*Direct solutions*

The dimension parameters of the PM are (unit: mm):

$a = 300$, $b = 150$, $d = 50$, $l_1 = 30$, $l_2 = 280$,

$l_3 = 140$, $l_4 = 180$, $l_5 = 90$, $l_6 = 230$.

Let the length of the connecting links 5 between the parallelograms and the length of the connecting link 8 are $l_7 = 0$ and $l_8 = 0$ respectively. At this point, the 3D model of the PM is shown in Fig. 4.

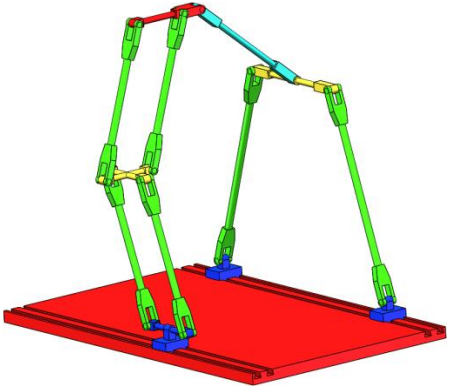

**FIGURE 4  3D CAD DESIGN**

The three input values $y_{A_1}$, $y_{A_2}$ and $y_{A_3}$ are:

$y_{A_1} = 162.6907$, $y_{A_2} = -143.3209$, $y_{A_3} = -24.6776$.

The direct solutions of the PM is calculated by MATLAB, as shown in Table 1.

**TABLE 1 THE VALUES OF DIRECT SOLUTIONS**

| No. | $x(mm)$ | $y(mm)$ | $z(mm)$ |
|---|---|---|---|
| 1 | 64.6353 | 175.6965 | 370.1818 |
| 2 | -128.8290 | 175.6965 | 119.7372 |
| 3* | -15.4714 | 9.6849 | 456.3315 |
| 4 | -128.8290 | 9.6849 | 118.2099 |

*Inverse solutions*

In Table 1, the direct solutions of the No. 3 group is substituted into the Eq. (21), and the 8 sets of inverse solution values of $y_{A_1}$, $y_{A_2}$ and $y_{A_3}$ are obtained, as shown in Table 2.

From Eqs. (18) and (19), there are theoretically 32 inverse solutions, but for this example, 24 solutions belong to the virtual number solutions, which does not exist for the real robot. Therefore only 8 numerical solutions exist.

**TABLE 2 THE VALUES OF INVERSE SOLUTIONS**

| No. | $y_{A_1}$ | $y_{A_2}$ | $y_{A_3}$ |
|---|---|---|---|
| 1 | 162.6909 | 22.6909 | 44.0476 |
| 2 | 162.6909 | 22.6909 | -24.6778 |
| 3 | 162.6909 | -143.3211 | 44.0476 |
| 4* | 162.6909 | -143.3211 | -24.6778 |
| 5 | -3.3211 | 22.6909 | 44.0476 |
| 6 | -3.3211 | 22.6909 | -24.6778 |
| 7 | -3.3211 | -143.3211 | 44.0476 |
| 8 | -3.3211 | -143.3211 | -24.6778 |

It can be seen that the inverse solutions data of the No. 4 group in Table 2 is consistent with the three input values given when the direct solution is solved, which proves the correctness of the direct and inverse solutions.

## SINGULARITY ANALYSIS

**Method of singularity analysis**

This paper uses the Jacobian matrix method to analyze the singularity configuration of the PM. Taking the first derivative of time $t$ from Eqs. (18) and (19), we have

$$\dot{\alpha} = -\frac{\dot{x}}{l_4 \sin \alpha} \quad (22)$$

$$\dot{\beta} = -\frac{\dot{x}}{l_6 \sin \beta} \quad (23)$$

Then, taking the first derivative of time $t$ from length constraint Eq. (20), and then substituting Eqs. (22) and (23) into the equations, there are

$$f_{i1} \dot{x} + f_{i2} \dot{y} + f_{i3} \dot{z} - u_{ii} \dot{y}_{A_i} = 0 \quad (i=1,2,3) \quad (24)$$

where

$u_{11} = y_{C_1} - y_{B_1}, u_{22} = y_{C_2} - y_{B_2}, u_{33} = y_{C_3} - y_{B_3}$.

$f_{11} = \cot\alpha(z_{C_1} - z_{B_1}), f_{12} = (y_{C_1} - y_{B_1}), f_{13} = (z_{C_1} - z_{B_1})$.
$f_{21} = \cot\alpha(z_{C_2} - z_{B_2}), f_{22} = (y_{C_2} - y_{B_2}), f_{23} = (z_{C_2} - z_{B_2})$.

$f_{31} = \cot\beta(z_{C_3} - z_{B_3}), f_{32} = (y_{C_3} - y_{B_3}), f_{33} = (z_{C_3} - z_{B_3}).$

Therefore, the relationship between the output speed $v_1 = \begin{bmatrix} \dot{x} & \dot{y} & \dot{z} \end{bmatrix}^T$ of the end effector and the actuated joint input speed $v_2 = \begin{bmatrix} \dot{y}_{A_1} & \dot{y}_{A_2} & \dot{y}_{A_3} \end{bmatrix}^T$ is

$$J_p v_1 = J_q v_2 \quad (25)$$

where

$$J_p = \begin{bmatrix} f_{11} & f_{12} & f_{13} \\ f_{21} & f_{22} & f_{23} \\ f_{31} & f_{32} & f_{33} \end{bmatrix} \quad J_q = \begin{bmatrix} u_{11} & & \\ & u_{22} & \\ & & u_{33} \end{bmatrix}$$

Whether the matrices $J_p$ and $J_q$ are singular, the singularity of the PM is divided into the following three categories:

(1) When $\det(J_q) = 0$, the PM has serial singularity.

(2) When $\det(J_p) = 0$, the PM has parallel singularity.

(3) When $\det(J_q) = \det(J_p) = 0$, the PM has comprehensive singularity.

**The result of singularity analysis**

(1) *Serial singularity*: When the PM has serial singularity, it means that each of the two branches near the drive link is folded or fully deployed. At this movement, $\det(J_q) = 0$, the DOF of the moving platform is reduced, and the set $W$ of the equation solution is:

$$W = \{W_1 \text{ Y} W_2 \text{ Y} W_3\} \quad (26)$$

where

$W_1 = \{y_{C_1} - y_{B_1} = 0\}$, that is, the three points $A_1$, $B_1$ and $C_1$ are collinear.

$W_2 = \{y_{C_2} - y_{B_2} = 0\}$, that is, the three points $A_2$, $B_2$ and $C_2$ are collinear.

$W_3 = \{y_{C_3} - y_{B_3} = 0\}$, that is, the three points $A_3$, $B_3$ and $C_3$ are collinear.

Example: The 3D configuration of satisfying condition $W_3$ is shown in Fig. 5.

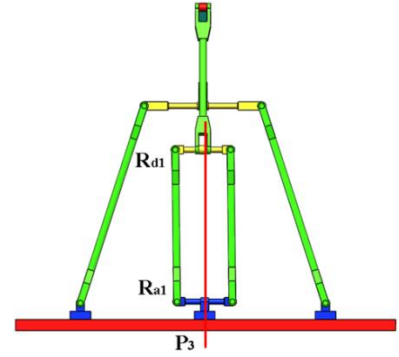

**FIGURE 5 EXAMPLE OF SERIAL SINGULARITY CONFIGURATION**

(2) *Parallel singularity*: When the PM has parallel singularity, it means that each branch is close to the link of the moving platform in a state of being folded together or fully deployed. At this movement, the DOF of the moving platform is increased, and even if the input link is locked, there may be has DOF output on the moving platform, assuming:

$$[f_{i1} \quad f_{i2} \quad f_{i3}] = \vec{e}_i \quad (i = 1,2,3)$$

If $\det(J_p) = 0$, the vectors $\vec{e}_1$, $\vec{e}_2$ and $\vec{e}_3$ have the following two cases:

(i) There are two vectors linear correlations

a) If $\vec{e}_1 = k\vec{e}_2$, that is, $[f_{11} \quad f_{12} \quad f_{13}] = k[f_{21} \quad f_{22} \quad f_{23}]$ is satisfied, the 3D configuration is that the vectors $\overrightarrow{B_1C_1}$ and $\overrightarrow{B_2C_2}$ are parallel in space, as shown in Fig. 6.

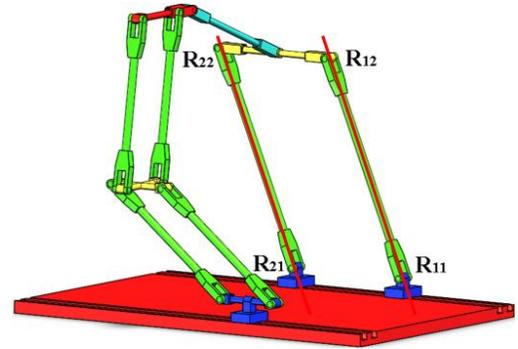

**FIGURE 6 EXAMPLE OF PARALLEL SINGULARITY CONFIGURATION**

b) If $\vec{e_1} = k\vec{e_3}$, that is, $[f_{11}\ f_{12}\ f_{13}] = k[f_{31}\ f_{32}\ f_{33}]$ is satisfied, there are:

$$\cot\alpha(z_{C_1} - z_{B_1}) = k\cot\beta(z_{C_3} - z_{B_3}),$$

$$(z_{C_1} - z_{B_1}) = k(z_{C_3} - z_{B_3}).$$

then, there is $\cot\alpha = \cot\beta$.

Due to the length of the link set by the PM, during the movement of the PM, there is always $\alpha \neq \beta$, that is, $\cot\alpha \neq \cot\beta$, then $\vec{e_1} \neq k\vec{e_3}$, similarly, $\vec{e_2} \neq k\vec{e_3}$.

(ii) There are three vectors linear correlations

If $\vec{e_2} = k_1\vec{e_1} + k_2\vec{e_3}\ (k_1 k_2 \neq 0)$, then, there is:

$$[f_{21}\ f_{22}\ f_{23}] = k_1[f_{11}\ f_{12}\ f_{13}] + k_2[f_{31}\ f_{32}\ f_{33}]$$

The calculation by MATLAB shows that the solution of $k_1$ and $k_2$ cannot be solved in this case, Therefore, this situation does not exist.

(3) *Comprehensive singularity*: $\det(J_q) = \det(J_p) = 0$, the serial singularity and the parallel singularity occur simultaneously. In this configuration, the actuated joints and the end effector of the PM have non-zero inputs and outputs that do not affect each other instantaneously, and the corresponding pose is the third kind of singularity. In this kind of singularity, the PM will lose the freedom and it should be avoided during the design phase of the PM.

**WORKSPACE ANALYSIS**

This paper uses the limit boundary search method to analyze the workspace of the PM, that is, the search range of the workspace is first assigned according to the length of the link. Then, based on the inverse position solution, all the points satisfying the constraint are searched, and the 3D map composed of these points is the workspace of the PM.

Determine the 3D search range of the workspace:

$-110 \leq x \leq 90$, $-250 \leq y \leq 250$, $180 \leq z \leq 480$ (unit: mm). The 3D workspace of the PM is obtained by MATLAB programming.

The serial singularity can be avoided by actual control, so this paper mainly discusses the parallel singularity trajectory. According to the link length constraint of Eq. (20) and the discriminant $\det(J_p) = 0$, the parallel singularity trajectory can be obtained, as shown in Fig. 7. Among them, the green part is the non-singularity workspace, and the red part is the singularity area, which indicates that there is a large non-singularity area inside the workspace. Fig. 8 is the projection view of the workspace in the XOZ and YOZ directions.

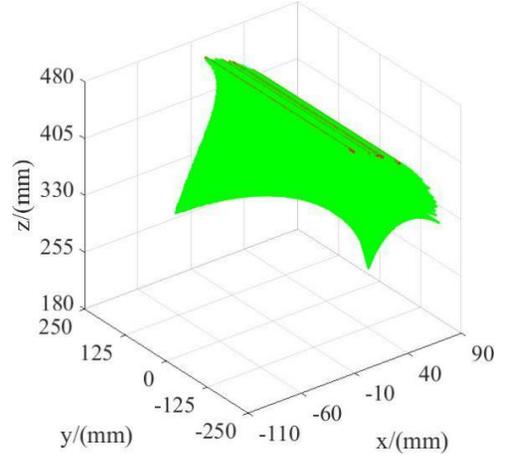

**FIGURE 7 WORKSPACE AND THE PARALLEL SINGULARITY SITUATION**

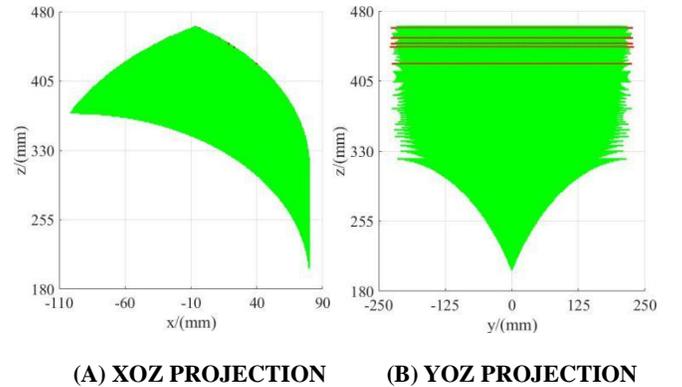

(A) XOZ PROJECTION   (B) YOZ PROJECTION

**FIGURE 8 PROJECTION VIEW OF THE WORKSPACE IN THE XOZ AND YOZ DIRECTION**

Fig. 9 shows four X-Y cross-sections in the Z direction of the workspace, which shows that the singularity and non-singularity workspace in each section also change with the change of Z value.

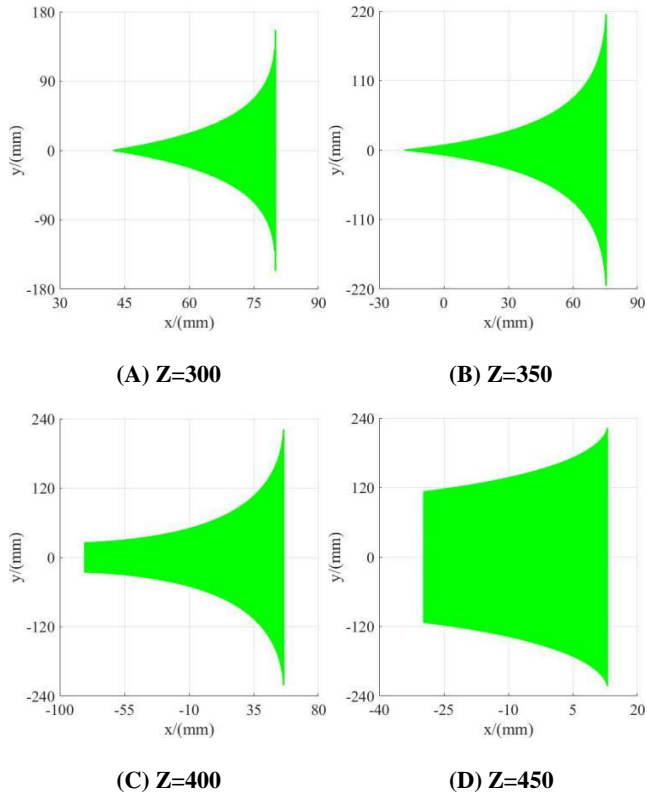

**(A)** Z=300     **(B)** Z=350

**(C)** Z=400     **(D)** Z=450

**FIGURE 9 DIFFERENT X-Y CROSS-SECTIONS IN THE WORKSPACE**

## CONCLUSIONS

The new 3-translational (3T) PM proposed in this paper has three advantages: (1) it is only composed of three actuated prismatic joints and other passive revolute joints, which is easy to be manufactured and assembled; (2) it has analytical direct position solutions, which brings the great convenience to error analysis, dimensional synthesis, stiffness and dynamics research; and (3) it has partial input-output motion decoupling, which is very beneficial to the trajectory planning and motion control of the PM.

According to the kinematics modeling principle proposed by the author based on the single-opened-chains method, in the first loop with positive constraint, the set one virtual variable $\gamma$ can be directly obtained by the special geometric constraint condition that the output link of the first loop always maintains the horizontal position (the condition is provided by the second loop with negative constraint). Therefore, the entire analytical position solutions are obtained without solving the virtual variable $\gamma$ by the geometric constraint equation in the second loop with negative constraint. This is the advantage of the topology of the PM being different from other PM, and it has analytical direct solutions. The method has clear physical meaning and simple calculation.

Based on the inverse solution, the conditions and locations of the three types of singularity configurations of the PM are obtained, and the size of the workspace of the PM and its parallel singularity area are given. The work of this paper lays the foundation for the stiffness, trajectory planning, motion control, dynamics analysis and prototype design of the PM.

## ACKNOWLEDGMENTS:

This research is sponsored by the NSFC (Grant No.51475050 and No.51375062) and Jiangsu Key Development Project (No.BE2015043).

## REFERENCES

[1] Clavel R. A Fast Robot with Parallel Geometry [C]. Proceeding of the 18th Int. Symposium on Industrial Robots. 1988: 91-100.

[2] Stock M, Miller K. Optimal Kinematic Design of Spatial Parallel Manipulators: Application to Linear Delta Robot [J]. Journal of Mechanical Design, 2003, 125 (2): 292-301.

[3] Bouri M, Clavel R. The Linear Delta: Developments and Applications [C]. Robotics. VDE, 2010: 1-8.

[4] Kelaiaia R, Company O, Zaatri A. Multiobjective optimization of a linear Delta parallel robot [J]. Mechanism & Machine Theory, 2012, 50 (2): 159–178.

[5] Tsai L. W., G. C. Walsh and R. E. Stamper, "Kinematics of a Novel Three DoF Translational Platform," IEEE International Conference on Robotics and Automation, Minneapolis, MN 3446-3451 (1996).

[6] Tsai L. W and S. Joshi, "Kinematics and optimization of a spatial 3-UPU parallel manipulator," ASME, J.Mech. Des. 122, 439-446 (2000).

[7] Zhao T., Huang Z., Kinematics analysis of a three dimensional mobile parallel platform mechanism. China Mechanical Engineering, 2001, 12 (6): 612-616.

[8] Yin X., Ma L.,Workspace Analysis of 3-DOF translational 3-RRC Parallel Mechanism [J]. China Mechanical Engineering, 2003, 14 (18): 1531-1533.

[9] X. Kong and C. M. Gosselin, "Kinematics and singularity analysis of a novel type of 3-CRR 3-DOFtranslational parallel manipulator," Int. J. Robot. Res. 21, 791-798 (2002).

[10] Li S, Huang Z, Zuo R. Kinematics of a Special 3-DOF 3-UPU Parallel Manipulator [C] ASME 2002 International Design Engineering Technical Conferences and Computers and Information in Engineering Conference. 2002: 1035-1040.

[11] Li S, Huang Z. Kinematic characteristics of a special 3-UPU parallel platform manipulator [J]. China Mechanical Engineering, 2005, 18 (3): 376-381.

[12] Yu J., Zhao T., Bi S., Comprehensive Research on 3-DOF Translational Parallel Mechanism. Progress in Natural Science, 2003, 13 (8): 843-849.

[13] Lu J, Gao G., Motion and Workspace Analysis of a Novel 3-Translational Parallel Mechanism [J]. Mechinery Design & Manufacture, 2007,11 (11): 163-165.

[14] Yang T., Topology Structure Design of Robot Mechanisms [M]. China Machine Press, 2004.

[15] Yang T., Liu A., Luo Y.,et.al，Theory and Application of Robot Mechanism Topology [M]. Science Press, 2012.

[16] Yang T., Liu A., Shen H. et.al, Topology Design of Robot Mechanism [M]. Springer, 2018.

[17] Yang T., Liu A., Shen H. et.al, Composition Principle Based on Single-Open-Chain Unit for General Spatial Mechanisms and Its Application,Journal of Mechanisms and Robotics，OCTOBER 2018, Vol. 10 / 051005-1~051005-16

[18] Zhao Y. Dimensional synthesis of a three translational degrees of freedom parallel robot while considering kinematic anisotropic property [J]. Robotics and Computer-Integrated Manufacturing, 2013, 29 (1): 169-179.

[19] Zeng Q, Ehmann K F, Cao J. Tri-pyramid Robot: Design and kinematic analysis of a 3-DOF translational parallel manipulator [M]. Pergamon Press, Inc. 2014.

[20] Zeng Q , Ehmann K F, Jian C. Tri-pyramid Robot: stiffness modeling of a 3-DOF translational parallel manipulator [J]. Robotica, 2016, 34 (2): 383-402.

[21] Lee S, Zeng Q, Ehmann K F. Error modeling for sensitivity analysis and calibration of the tri-pyramid parallel robot [J]. International Journal of Advanced Manufacturing Technology, 2017 (5): 1-14.

[22] PRAUSE I, CHARAF E. Comparison of Parallel Kinematic Machines with Three Translational Degrees of Freedom and Linear Actuation [J]. CHINESE JOURNAL OF MECHANICAL ENGINEERING, Vol. 28, No.4, 2015.

[23] Mahmood M, Mostafa T. Kinematic Analysis and Design of a 3-DOF Translational Parallel Robot [J]. International Journal of Automation and Computing, 14 (4), August 2017, 432-441.

[24] Shen H., Xiong K., Meng Q., Kinematic decoupling design method and application of parallel mechanism [J] .Transactions of The Chinese Society of Agricultural Machinery, 2016, 47 (6): 348-356.